\documentclass[journal]{IEEEtran}
\pdfoutput=1

\usepackage{cite}
\usepackage[cmex10]{amsmath}
\usepackage{units}
\usepackage{lipsum}
\usepackage[caption=false]{subfig}
\usepackage{graphicx}	
\usepackage{pstricks}
\usepackage{physics}
\usepackage{float}
\usepackage{eurosym}
\usepackage{multicol}
\usepackage{booktabs}
\usepackage{units}
\usepackage{amsmath,amssymb}
\usepackage{pifont}
\usepackage{soul}
\usepackage{bm}

\renewcommand{\theta}{\vartheta}

\graphicspath{{Figure/}}
\begin{document}

\title{NMPC trajectory planner for urban autonomous driving}


\author{
    \IEEEauthorblockN{F. Micheli\IEEEauthorrefmark{1}, M. Bersani\IEEEauthorrefmark{2}, S. Arrigoni\IEEEauthorrefmark{2}, F. Braghin\IEEEauthorrefmark{2}, F. Cheli\IEEEauthorrefmark{2}}\\
    \IEEEauthorblockA{\IEEEauthorrefmark{1}ETH Z\"urich, Automatic Control Laboratory, Z\"urich, Switzerland}\\
    \IEEEauthorblockA{\IEEEauthorrefmark{2}Politecnico di Milano, Department of Mechanical Engineering, Milano, Italy}
}

\maketitle

\begin{abstract}

This paper presents a trajectory planner for autonomous driving based on a Nonlinear Model Predictive Control (NMPC) algorithm that accounts for Pacejka’s nonlinear lateral tyre dynamics as well as for zero speed conditions through a novel slip angles calculation. In the NMPC framework, road boundaries and obstacles (both static and moving) are taken into account thanks to soft and hard constraints implementation. 	The numerical solution of the NMPC problem is carried out using ACADO toolkit coupled with the quadratic programming solver qpOASES. The effectiveness of the proposed NMPC trajectory planner has been tested using CarMaker multibody models. 
Time analysis results provided by the simulations shown, state that the proposed algorithm can be implemented on the real-time control framework of an autonomous vehicle under the assumption of data coming from an upstream estimation block. 

\end{abstract}

\IEEEpeerreviewmaketitle

\section{introduction}
Autonomous driving is considered one of the most disruptive technologies of the near future that will completely reshape transportation systems as well as, hopefully, its impact on safety and environment \cite{harper2016estimating,greenblatt2015automated}.

According to \cite{paden2016survey} and \cite{li2017development}, a typical control architecture for AVs is composed by four hierarchical control layers called: route planning layer, behavioural layer, trajectory planning layer, local feedback layer.

Focusing on the trajectory planning task, different algorithms can be found in literature. Potential field algorithms consider the ego-vehicle as an electric charge that moves inside a potential field~\cite{khatib1985real}. The final desired configuration is represented by an attractive field, while obstacles are represented by repulsive fields. Extensions of this method are the virtual force fields~\cite{borenstein1989real} and the vector field histogram algorithms~\cite{Borenstein1991}. The implementation is simple but the solution usually gets trapped in local minima~\cite{koren1991potential}.

Graph search based algorithms perform a global search on a graph generated by discretization of the path configuration space.
These methods do not explicitly enforce dynamic constraints, but explore the reachable and free configuration space using some discretization strategy, while checking for steering and collision constraints. In order to obtain smooth paths, several sampling-based discretization strategies are available in literature, such as cubic splines~\cite{judd2001spline} or clothoids~\cite{fleury1995primitives}. The main drawback of these methods is related to the discretization process itself that prevents to obtain a true optimal path in the configuration space.

Incremental search techniques, such as the Rapidly-exploring Random Tree (RRT) algorithm \cite{lavalle1998rapidly} and its successful asymptotically optimal adaptation, the RRT* algorithm \cite{karaman2010optimal}, try to overcome the limitations of graph search methods by incrementally producing an increasingly finer discretization of the configuration space. The use of RRT algorithms in combination with dynamic vehicle models and in presence of dynamic obstacles while maintaining real-time performances remains an active research area~\cite{berntorp2017path,pepy2006path}.

Recently, particular attention has been devoted to path-velocity decomposition techniques. The trajectory planning task is decomposed into two sub-tasks: path planning and velocity planning. Various methods are available to generate kinematically feasible paths, such as cubic curvature polynomials~\cite{nagy2001trajectory}, Bezier curves~\cite{
	chen2014quartic}, clothoid tentacles~\cite{alia2015local,
	de2009flatness} methods. The velocity planning, instead, can be performed either assuming a speed profile with a certain predefined shape or using other techniques, such as MPC, to obtain the optimal velocity profile along the chosen path~\cite{qian2016motion}. The main drawback in path-velocity decomposition approaches lies in the handling of dynamic obstacles: due to the separation of path and velocity planning tasks, it is difficult to account for the time-dependent position of dynamic obstacles~\cite{qian2016motion}.

In the last decades, progress in theory and algorithms allowed to expand the use of real-time LMPC and NMPC to the area of autonomous vehicles for tracking tasks \cite{falcone2007predictive,falcone2008low} and, more recently, for trajectory generation tasks~\cite{liniger2015optimization,frasch2013auto}. The use of LMPC and NMPC for trajectory generation and path tracking has been particularly successful due to its ability to handle Multi-Input Multi-Output (MIMO) systems, while considering vehicle dynamics and constraints on state and input variables. The various approaches available in literature differ mainly for the complexity of the vehicle model, the presence of static and/or dynamic obstacles and the optimization method used to solve each MPC iteration.

A common approach to reduce the computational burden of trajectory generation and tracking tasks is to decompose the problem into two subsequent hierarchical MPCs: the higher level controller adopts a simple vehicle model in order to plan a trajectory over a long time horizon while the lower level controller tracks such trajectory using a more complex vehicle model over a shorter horizon.

The main risk is that, due to unmodelled dynamics, the high level controller can generate trajectories that cannot be tracked by the low level controller. A possible solution has been presented in \cite{gray2012predictive} where the high level NMPC is based on precomputed primitive trajectories that results advantageous especially in structured environments.
The drawback is that the high level controller is allowed to choose the (sub-)optimal ``manoeuvre'' only from a finite set of precomputed primitive trajectories. Furthermore, the real-time implementation results challenging due to the necessity of solving an on-line mixed-integer optimization problem or of managing a large precomputed look-up table.
 
In \cite{gao2012spatial} a two-level NMPC that uses a nonlinear dynamic single-track vehicle model for the high level trajectory generation and a more complex four wheel vehicle model for the low level trajectory tracking is proposed. The single-track vehicle model uses a simplified Pacejka's tyre model and exploits a spatial reformulation to speed up calculations and defines constraints as a function of the spatial coordinates. The controller is able to avoid static obstacles and the use of a dynamic model greatly improves the ability of the low level controller to follow the prescribed trajectory.

The same spatial reformulation has been implemented in \cite{frasch2013auto} where the tracking infeasibility issue of two-levels schemes has been solved by adopting a single level NMPC based on a complex model to merge the trajectory generation and tracking tasks. 

The spatial reformulation adopted in \cite{gao2012spatial} and \cite{frasch2013auto} implies some important limitations, i.e. the vehicle's velocity must always be different from zero and the integration is performed in space. Furthermore, the advantages of the spatial reformulation are greatly reduced when dynamic obstacles are considered.

In \cite{liniger2015optimization} the trajectory planning is performed using a single stage NMPC. In this case a single-track vehicle model with Pacejka's lateral tyre model is adopted and the model is linearized at each step around the current configuration to speed up calculations. Moving obstacles are considered as fixed within each MPC step, thus reducing the prediction capability of the controller.

A single layer linear time-varying MPC based on a kinematic vehicle model is implemented in \cite{gutjahr2017lateral}. Static and moving obstacles are considered, and the optimal control problem (OCP) is solved by using a slack formulation and applying a quadratic programming (QP) routine.

The use of kinematic and dynamic single-track vehicle models for MPC applications is investigated in \cite{kong2015kinematic}. On one hand, kinematic models have the advantage of a lower computational cost and do not suffer from the slip going to infinity at zero velocity. On the other hand, kinematic models are not able to consider friction limits at tyre-road interaction. Hence, the paper concludes suggesting the use of dynamic models when moderate driving speeds are involved.

As previously proposed in \cite{inghilterra2018,trabalzini2019,Arrigoni2021mpc} by the authors, this paper presents a single stage NMPC trajectory planning algorithm that can deal with multiple static and moving obstacles. The vehicle model is implemented as a single-track vehicle model in which lateral tyre dynamics is modelled through Pacejka's simplified Magic Formula model \cite{pacejka1992magic}. A novel modified slip calculation is introduced allowing to avoid the need of switching to a different model in low speeds and stop-and-go scenarios that are common in urban driving. The algorithm is implemented using ACADO toolkit \cite{houska2011acado} coupled with the QP solver qpOASES \cite{ferreau2014qpoases}. For validation purposes, the proposed trajectory planner has been tested on two relevant urban driving scenarios in Simulink 
and CarMaker \cite{carmaker2014ipg} co-simulation environment.

\section{The NMPC Framework}\label{section2}
The aim of the modelling phase is to find the best trade-off between model complexity, accuracy and computational cost. On one hand, the implementation of a good mathematical model ensures accurate predictions of the system's behavior. On the other hand, if the computational effort raises too much, real-time applicability is not achievable. Moreover, higher levels of detail usually entail an often-difficult parameter estimation step that can lead to a lower overall robustness.

\subsection{Road Map Model}\label{sec:RoadMapModel}
A reliable local road map is fundamental for the navigation of AVs and for the prediction of their position with respect to obstacles and to road boundaries along the optimization horizon. Therefore, a local road map based has been developed in the local reference framework $s$-$y$ (curvilinear abscissa and lateral displacement) as depicted in Fig.~\ref{Fig-Vehicle_model}).
This framework is particularly convenient when describing road boundaries and obstacles' motion along the prediction horizon. Starting from a road map defined in the Cartesian global reference system $X$-$Y$, the local road map is generated by fitting third-order polynomials that locally approximate the road centreline angle $\theta_c$ 
as a function of the curvilinear abscissa $s$. 

Throughout this work, the complete knowledge of the road map is assumed. Thus, all the polynomial coefficients are available to the trajectory planner as a function of the vehicle's current position along the map. As stated in \cite{cinesi_curva}, considering the availability of a detailed road map is a reasonable assumption when dealing with the control of autonomous vehicles.This information can, for example, be obtained as the of an on-line environment mapping process based on vehicle mounted sensors such as cameras, lidars and radars~\cite{Mentasti2019, kocic2018sensors}.

\subsection{Vehicle Model} \label{VehicleModel}
The vehicle model adopted in this work is a single-track dynamic model in which the lateral tyres forces are calculated through a simplified Pacejika's model based on a modified slip calculation procedure. This vehicle model represents a trade-off between the high complexity of 14 (or more) dofs models \cite{braghin2009dinamica,shim2007understanding} and single-track kinematic models \cite{kong2015kinematic,rajamani2011vehicle}, allowing to reduce the number of state variables while capturing the relevant nonlinearities associated to lateral tyre dynamics.

\begin{figure}
	\centering
	{\includegraphics[width=0.32\textwidth]{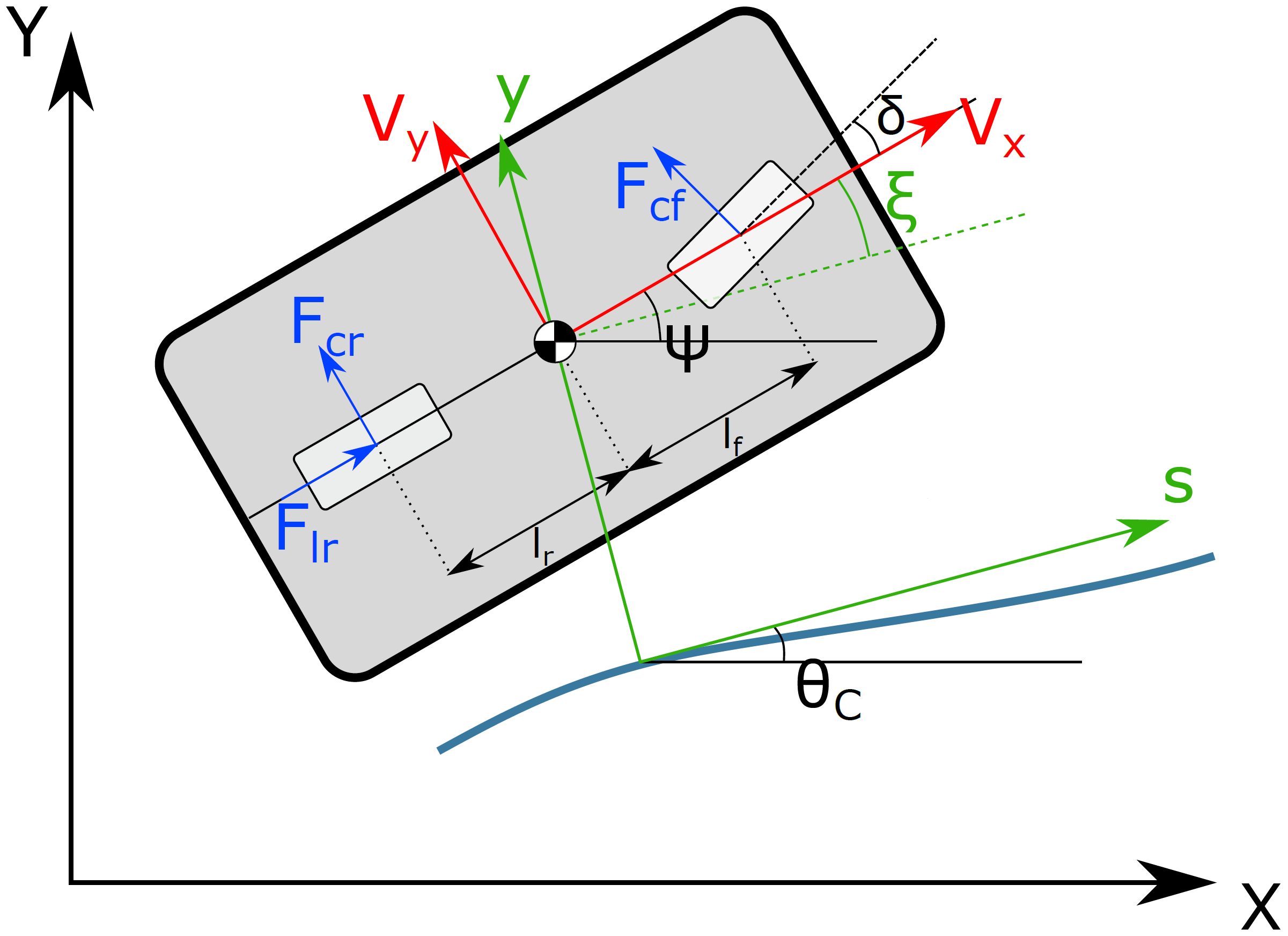}}
	\caption{Representation of the vehicle model framework. $X$-$Y$ represents the global reference system, while $s$-$y$ denotes the curvilinear abscissa reference system. $V_x$-$V_y$ are oriented according to the chassis' reference system.}
	\label{Fig-Vehicle_model}
\end{figure}

The model considers two translational dofs in the $X$-$Y$ plane and one rotational dof around the $Z$ axis as shown in Fig.~\ref{Fig-Vehicle_model}. The translation along the Z direction and the roll and pitch rotations are neglected. The system's equations of motion in state-space form $\dot{\bm{x}}(t)=f(\bm{x}(t),\bm{u}(t))$, expressed in the local reference system $s$-$y$ (see \cite{NILSSON1984} and \cite{micaelli1993trajectory}) are reported in \eqref{eq:SingleTrack_sy}, where $\bm{x}(t)\in\mathbb{R}^8$ is the state vector and $\bm{u}(t)\in\mathbb{R}^2$ is the input vector.

\begin{equation}\label{eq:SingleTrack_sy}
\left\{ \begin{aligned}
\dot{s}\ &=\frac{V_x\cos(\xi)-V_y\sin(\xi)}{1-\theta_{cs}'y}\\
\dot{y}\ &=V_x\sin(\xi)+V_y\cos(\xi)\\
\dot{\xi}\ &=\omega -\theta_{cs}'\dot{s}\\
\dot{V_x}&=\omega V_y+\frac{F_{rl}}{M} - \frac{F_{fc}\sin(\delta)}{M} - \frac{F_{aero}}{M}\\
\dot{V_y}&=-\omega V_x + \frac{F_{rc}}{M} + \frac{F_{fc}\cos(\delta)}{M}\\
\dot{\omega}\ &=\frac{1}{J_z}(-F_{rc}l_r+F_{fc}l_f\cos(\delta))\\
\dot{\delta}\ &=u_1\\
\dot{T_r}&=u_2
\end{aligned} \right.
\end{equation}

\noindent with $s$ is the curvilinear abscissa, $y$ is the distance between the vehicle's centre of mass (C.o.M.) and the centreline, $\xi$ is the error between the heading angle of the vehicle and the angle of the tangent to the road centreline in the global reference system ($\xi=\psi - \theta_c$), $V_x$, $V_y$ and $\omega$ are the vehicle's velocity components in the local reference system centred in the C.o.M., $\delta$ and $T_r$ are the steering angle at the front wheels and the torque applied to the rear axle and are obtained by integrating the inputs $u_1$ and $u_2$. 

The longitudinal force $F_{rl}$ at the rear axle is equal to $F_{rl}=T_r/R_r$, with $T_r$ being the total applied torque and $R_r$ the nominal rear wheels radius. This assumption is acceptable in (relatively) low speed urban scenarios and avoids introducing additional dofs to describe the wheels' rotation.

The lateral forces $F_{fc}$ and $F_{rc}$ at the front and rear tyres are calculated using the well-known Pacejka's semi-empirical tyre model \cite{pacejka1992magic}, also known as \textit{``Magic Formula''}, considering steady-state pure lateral slip conditions:

\begin{small}
\begin{equation*}\label{eq:PacejkaMagic}
F_{*c}=-2D\sin(C\atan(B\alpha_*+E(\atan(B\alpha_*)-B\alpha_*)))
\end{equation*}
\end{small}

\noindent where the pre-multiplication by $2$ is used to take into account that there are two tyres for each axle, $B$, $C$, $D$, $E$ are the Pacejka's coefficients and $\alpha_*$ is the slip angle of the front or rear tyres that are calculated as:

\begin{equation}\label{eq:PacejkaSlips}
\begin{aligned}
\alpha_f &= \atan\left( \frac{(V_y+\omega\, l_f)\cos(\delta)-V_x\sin(\delta)} {V_x\cos(\delta)+(V_y+\omega\, l_f)\sin(\delta)} \right)\\
\alpha_r &= \atan\left( \frac{V_y-\omega\, l_r} {V_x} \right) 
\end{aligned}
\end{equation}

The coupling of the lateral and longitudinal tyre-road contact forces is indirectly accounted for by considering the coupling of the lateral and longitudinal accelerations through the equivalent friction ellipse constraint~\cite{brach2000modeling}:
\begin{small}
\begin{equation}\label{eq:ellisse_Tyre}
\begin{aligned}
T_r\leq R_rM\Bigg[ a_1 \sqrt{1-\! \frac{\dot{V}_y^2}{b_1^2}}-\! \Big[\omega V_y - \frac{F_{fc}\sin(\delta)}{M} - \frac{F_{aero}}{M}\Big] \Bigg]\\
T_r\geq R_rM\Bigg[\! -a_2 \sqrt{1-\! \frac{\dot{V}_y^2}{b_2^2}}-\! \Big[\omega V_y - \frac{F_{fc}\sin(\delta)}{M} - \frac{F_{aero}}{M}\Big] \Bigg]
\end{aligned} 
\end{equation}
\end{small}

\noindent where coefficients $a_*$ and $b_*$ are fitted to approximate the ellipse-shaped envelope as a function of the estimated friction coefficient and road surface conditions. 
\\Through \eqref{eq:ellisse_Tyre} the maximum longitudinal acceleration (i.e. torque applied to the tyres) is a function of the lateral acceleration of the vehicle's C.o.M. This approach is acceptable in urban driving conditions (not in limit handling conditions) which in fact is the target for the proposed trajectory planner. Moreover, this approach allows to easily enforce safety: as proposed in~\cite{brach2011tire}, the friction ellipse approximation can be artificially reduced with respect to the real tyre-road friction limit to always maintain a safety margin.

\begin{figure}
	\centering
	{\includegraphics[width=0.3\textwidth]{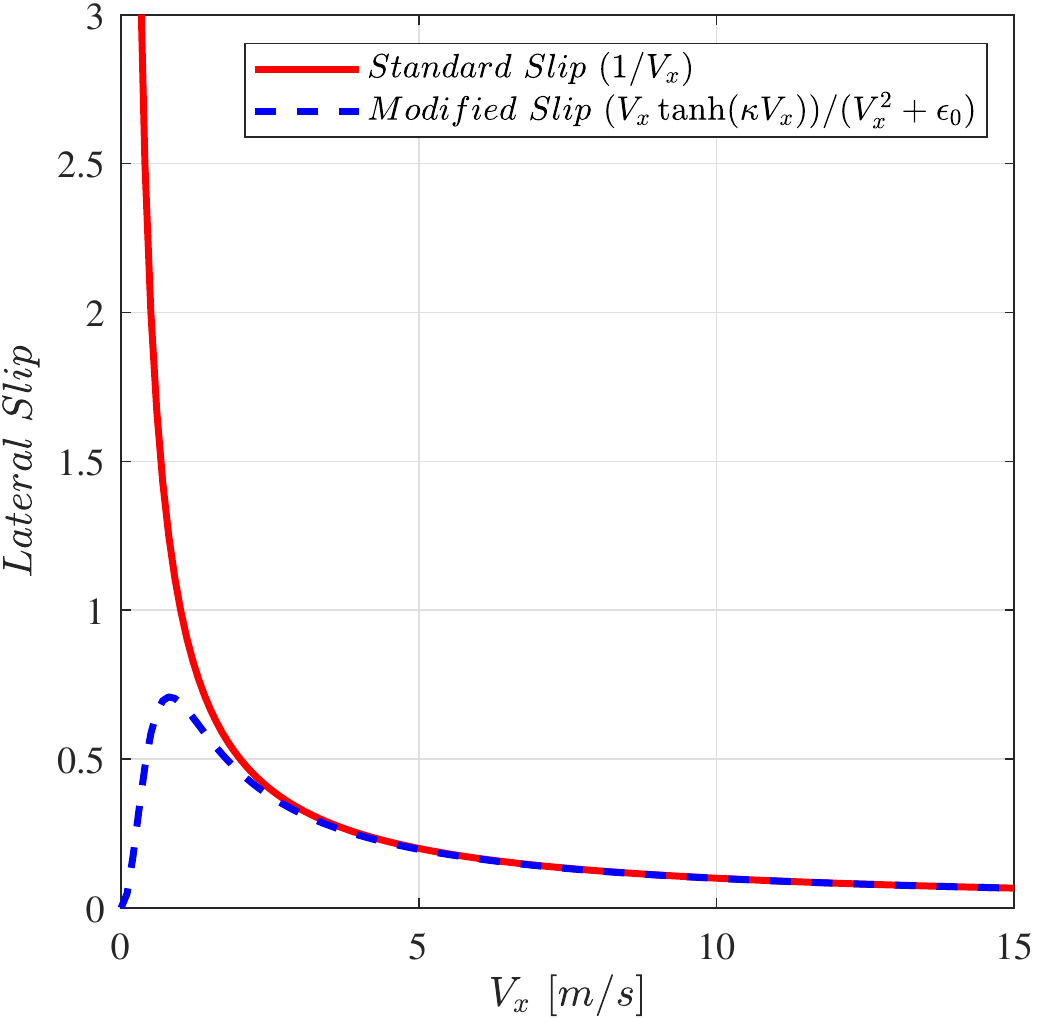}}
	\caption{Comparison between the standard and the modified slip curves with $\kappa=2$ and $\epsilon_0=0.4$.}
	\label{fig:Slip_Mod_Comparison}
\end{figure}

It is important to point out that equations \eqref{eq:PacejkaSlips} can be difficult to handle when $V_x \to 0\ m/s$. In fact, as $V_x$ approaches zero, the slips tend to infinity for non-vanishing $V_y$ or $\omega$. This causes model instability even for small numerical or estimation errors.

Since the intent is to develop a trajectory planner for urban environment, it is necessary to have a model that is robust and stable also at low or zero velocities. A velocity dependent switch to a kinematic vehicle model would introduce additional computational costs and should therefore be avoided. Thus, the slip angle calculation has been modified as
\begin{align}
\alpha_f\! &=\! \atan\! \left(\! \frac{[(V_y+\omega\, l_f)\cos(\delta)\!-\!V_x\sin(\delta)]\,  V_x \tanh(\kappa V_x)} {[V_x\cos(\delta)+(V_y+\omega\, l_f)\sin(\delta)]\, V_x+\epsilon_0} \! \right)\notag \\
\alpha_r\! &=\! \atan\left(\! \frac{[V_y-\omega\, l_r]\, V_x\tanh(\kappa V_x)} {V_x^2+\epsilon_0} \! \right)\label{eq:PacejkaSlips-mod}
\end{align}

\noindent where $\kappa$ and $\epsilon_0$ allow to shape the transition toward a kinematic-like behaviour at low velocities while preserving the Pacejka's model slip calculation at medium-high speeds as shown in Fig.~\ref{fig:Slip_Mod_Comparison}.

\begin{figure}
	\centering
	{\includegraphics[width=0.3\textwidth]{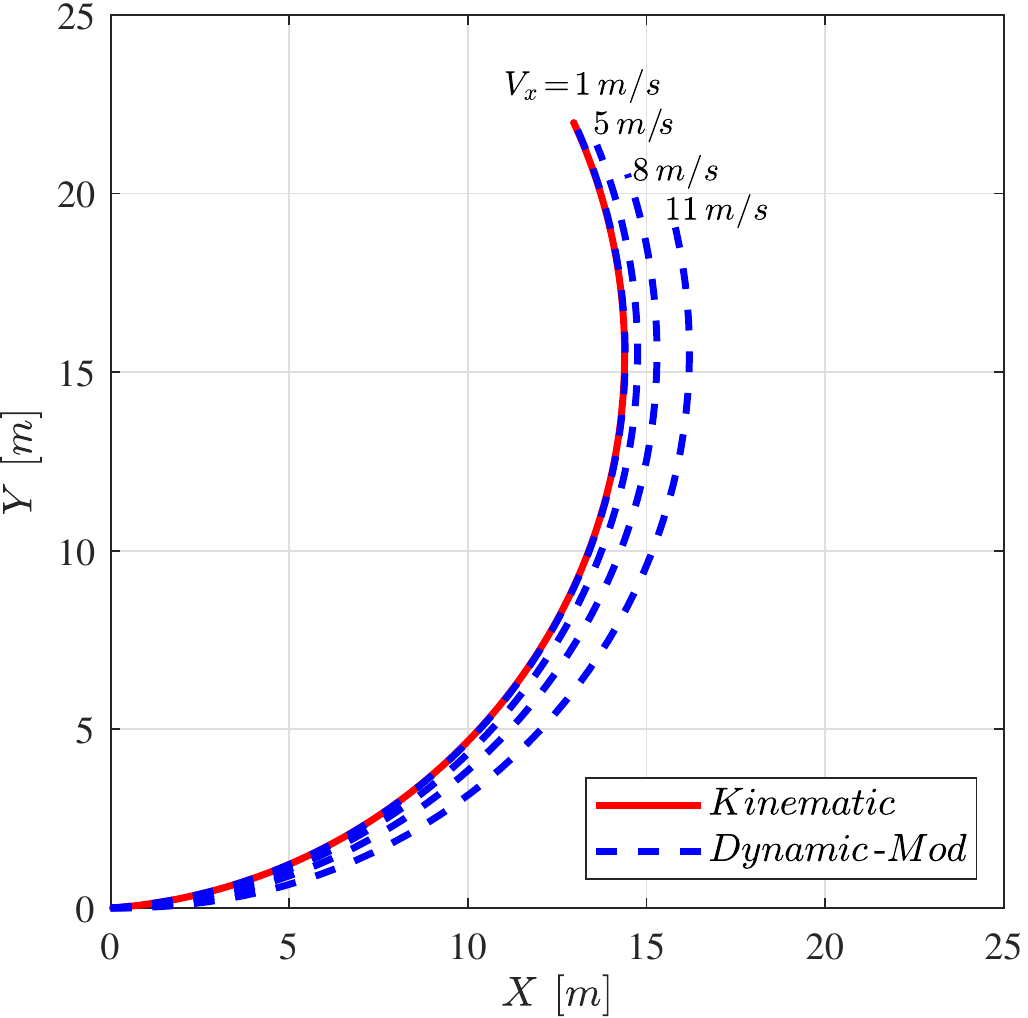}}
	\caption{Comparison between the dynamic and the kinematic vehicle models C.o.M. trajectories obtained imposing a fixed reference steering input $\delta = 0.15\ rad$ at the front wheels and constant longitudinal velocities $V_x = \{1,5,8,11\}\,m/s$.}
	\label{fig:Kinematic_Dynamic}
\end{figure}

To assess the effectiveness of the modified slip calculation at low velocities, the dynamic vehicle model behaviour at different velocities is compared to the single-track kinematic vehicle model \cite{kong2015kinematic,rajamani2011vehicle}. In Fig. \ref{fig:Kinematic_Dynamic} are shown the vehicle C.o.M. trajectories obtained imposing a fixed reference steering input $\delta = 0.15\ rad$ at the front wheels and constant longitudinal velocities $V_x = \{1,5,8,11\}\,m/s$. The effect of the lateral slip is clearly visible as the longitudinal velocity increases, while at (very) low velocities the dynamic model behaves like the kinematic one.

\begin{figure}
	\centering
	{\includegraphics[width=0.3\textwidth]{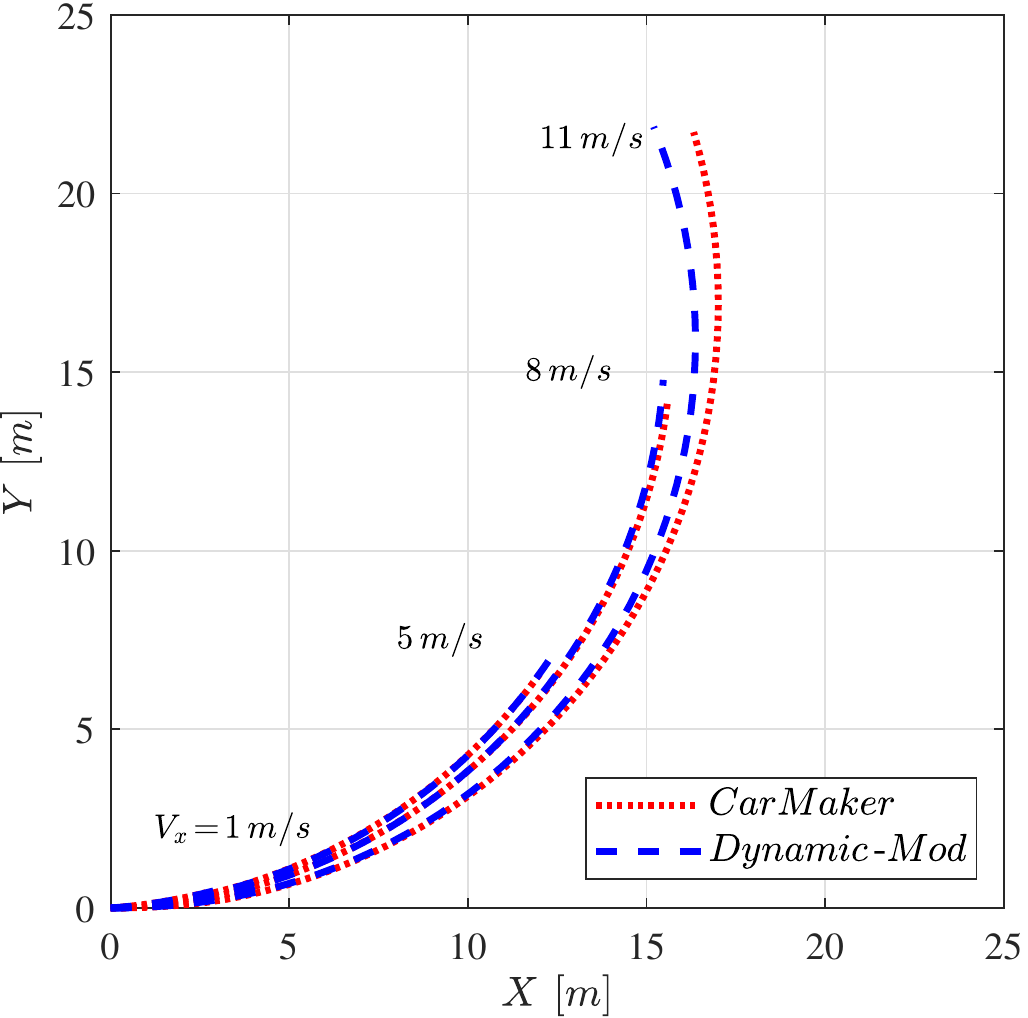}}
	\caption{Comparison between CarMaker $18$ dofs model and the dynamic single-track model with modified slip calculation C.o.M. trajectories for a fixed reference steering input $\delta = 0.15\ rad$ at the front wheels and constant longitudinal velocities $V_x = \{1,5,8,11\}\,m/s$.}
	\label{fig:SingleTrack_CarMaker}
\end{figure}

The single-track vehicle model with modified slip angle calculation has been also compared with an 18 dofs multibody vehicle model implemented in CarMaker environment. The results of the simulations are shown in Fig.~\ref{fig:SingleTrack_CarMaker} where both models are subjected to a constant steering input $\delta = 0.15\ rad$ at the front wheels and constant longitudinal velocities $V_x = \{1,5,8,11\}\,m/s$ for $3\,s$.
At low velocities, the simplified 3 dofs model moves on trajectories that are very close to the ones of the more complex CarMaker model. As velocity increases, the single-track vehicle model predicts tighter curvature radii. This is due to the fact that the 3 dofs model does not account for the load transfers that cause a reduction of the tyre cornering stiffness~\cite{mastinu2014road}. The single-track vehicle trajectory at $11\ m/s$ has a curvature radius of $16.4\ m$, while a radius of $17.0\ m$ is obtained using the CarMaker model. It should be noted that this represents a pretty tough manoeuvre, that exploits approximately $73\%$ of the tyre friction limit.

Since the trajectory planner has been designed for low to medium velocities in urban driving scenarios, the single-track vehicle model~\eqref{eq:SingleTrack_sy} with the modified slip calculation \eqref{eq:PacejkaSlips-mod} represents a good approximation of the complex CarMaker model. Furthermore, differences between the estimated and the actual vehicle trajectories can be addressed by the feedback provided by the NMPC logic that, at each step, re-plans the optimal trajectory, compensating for small modelling errors and external disturbances.

\subsection{Constraints}
The OCP is subjected to both hard and soft constraints. In fact, soft constraints allow to obtain a smoother driving behaviour while improving the convergence of the solver and reducing the risk of hard constraints violations in presence of modelling errors, external disturbances and uncertainties (eventually coming from sensors) related to obstacles’ current and future estimated positions.

\subsubsection*{Constraints on state and input variables}
this set of constraints includes the physical limits of the vehicle model:
\begin{small}
\begin{equation}\label{eq:PhysicalConstraints}
\left\{ \begin{aligned}
\delta_{min} &\leq \, \delta \, \leq \delta_{max}& \textit{(steering angle [rad])},\\
{T_r}_{min} &\leq {T_r} \leq {T_r}_{max}& \textit{(torque [Nm])},\\
\dot{\delta}_{min} &\leq \, \dot{\delta} \, \leq \dot{\delta}_{max}& \textit{(steering angular velocity [rad/s])},\\
\dot{T_r}_{min} &\leq \dot{T_r} \leq \dot{T_r}_{max}& \textit{(torque derivative [Nm/s])}.\\
\end{aligned}
\right.
\end{equation}
\end{small}

Physical limits have been derived from the experimental vehicle presented in~\cite{vignati2018autonomous,vignati2018transform} and they have been applied to both the simulated and the CarMaker multibody vehicle models as constraints. In this way, simulation results are already fitted on the real vehicle that, in a next phase of this work, will become the prototype on which this planner will be implemented.\\
The steering angle $\delta$ is limited by the geometry of the steering system while torque $T_r$ is bounded by the friction limit and the maximum torque that can be generated by the powertrain. The steering angular velocity $\dot{\delta}$ and the torque derivative $\dot{T_r}$ are limited according to the saturation limits of the actuators installed on the prototype vehicle. These constraints are handled by the active-set routine of the QP solver qpOASES~\cite{ferreau2014qpoases}.

\subsubsection*{Presence of obstacles}
both static or moving obstacles (such as cars, trucks and motorcycles driving along the road as well as pedestrians crossing the road, vehicles parked on the road side, etc.) are taken into account. In case of moving obstacles, it is fundamental to estimate not only the obstacle's current position, but also its future trajectory. The future obstacle's positions are approximately obtained in the local reference frame $s$-$y$ assuming constant velocities $V_s^{obs}$ and $V_y^{obs}$ in $s$ and $y$ directions (simplest possible assumption) as in \eqref{eq:Constraints_ObstMotion}. The adoption of the local $s$-$y$ reference frame greatly simplifies the description of the obstacles' motion with respect to a more standard Cartesian reference system: two parameters are sufficient to reasonably approximate obstacles' motion even along twisty roads.
\begin{equation}\label{eq:Constraints_ObstMotion}
\left\{ 
\begin{aligned}
s^{obs}&=s_0^{obs}+V_s^{obs}\, t\\
y^{obs}&=y_0^{obs}+V_y^{obs}\, t
\end{aligned}
\right.
\end{equation}

Since equations \eqref{eq:Constraints_ObstMotion} are time dependent, the equation $\dot{t}=1$ has been added to \eqref{eq:SingleTrack_sy} to integrate time along the prediction horizon. Note that the trajectory planner requires the knowledge of the obstacles' positions and velocities at each time sample. This information could be obtained from sensors installed on the vehicle (such as cameras, lidars and radars)~\cite{Mentasti2019,kocic2018sensors}.

\begin{figure}
	\centering
	{\includegraphics[width=0.2\textwidth]{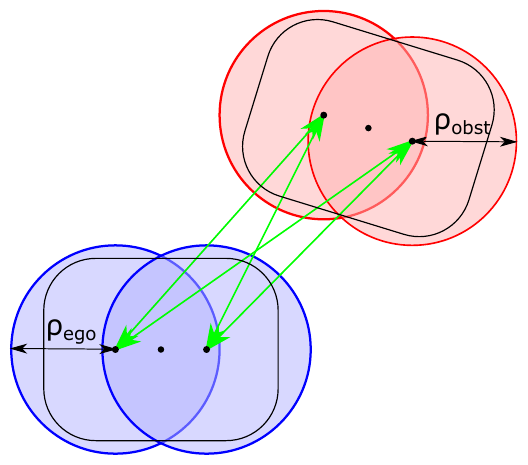}}
	\caption{Schematic representation of ego-vehicle and obstacle by means of multiple circles.}
	\label{fig:Ostacoli_CerchioCerchio}
\end{figure}

One of the simplest yet effective ways to model the ego-vehicle and the obstacles is by means of multiple circles~\cite{gutjahr2017lateral}. In this way, the relative distance calculation can be approximated by an Euclidean norm in the $s$-$y$ space. If the ego-vehicle and the obstacle are both modelled through two circles, as shown in Fig. \ref{fig:Ostacoli_CerchioCerchio}, a total of four distance computations is required to evaluate each obstacle constraint.

Each constraint is implemented both as a hard \eqref{eq:ConstraintsOst-DoppioCerchio-hard} and as a soft \eqref{eq:ConstraintsOst-DoppioCerchio-soft} constraint:
\begin{align}
&(s^{ego}_{c,\star}-s^{obs}_{c,\star})^2+(y^{ego}_{c,\star}-y^{obs}_{c,\star})^2\geq (\rho^{ego}+\rho^{obs})^2\label{eq:ConstraintsOst-DoppioCerchio-hard}\\
&C_{obs} = e^{\ \kappa\big[(\rho^{ego}+\rho^{obs})^2 - (s^{ego}_{c,\star}-s^{obs}_{c,\star})^2-(y^{ego}_{c,\star}-y^{obs}_{c,\star})^2 \big]}\label{eq:ConstraintsOst-DoppioCerchio-soft}
\end{align}

\noindent where $s_{c,\star}$ and $y_{c,\star}$ (with $(\star)\in \{1, 2\}$) represent the position of the centre of the circles of radii $\rho^{ego}$ and $\rho^{obs}$ considering the ego-vehicle heading $\psi$ and approximating the obstacle constant heading as $\psi^{obs}=\atan (V_y^{obs}/V_s^{obs})$.

An extra soft constraint can be added as an extra penalty cost to enforce the minimum safety distance:
\begin{equation}\label{eq:ConstraintsOst-Ellissoide-soft}
C_{safety} = e^{\ \kappa\big[ -d_s(s^{ego}-s^{obs})^2 -d_y(y^{ego}-y^{obs})^2 \big]}
\end{equation}
This constraint is shaped as an ellipse in the $s$-$y$ reference frame to guarantee the safety distance also during curves. 
The ellipse is centred with respect to the obstacle while its dimensions (i.e. parameters $d_s$ and $d_y$) are set by a higher level behavioural layer as a function of the ego-vehicle velocity, traffic conditions and maximum braking capability that usually depends on weather and road surface conditions.

Thanks to equation~\eqref{eq:ConstraintsOst-Ellissoide-soft}, also the uncertainty related to the obstacles' predicted velocities can be accounted for. In fact, while the obstacles' actual initial position estimation can be considered as being highly accurate (depending on the reliability of the sensors and of the estimation algorithms), the reliability of the prediction of future obstacles' positions relies on the assumption of constant velocities $V_s^{obs}$ and $V_y^{obs}$. This uncertainty can be accounted for by reducing the weights $d_s$ and $d_y$ in equation \eqref{eq:ConstraintsOst-Ellissoide-soft}, thus increasing the ellipse dimensions (e.g. as a function of time).

\subsubsection*{Road boundaries}
the distance $y$ of the road boundaries from the centreline as a function of the curvilinear abscissa $s$ is described through third order polynomials that are considered to be known as explained in Section~\ref{sec:RoadMapModel}. Similarly as discussed in the previous paragraph , the vehicle can be approximated with two circles of radius $\rho^{ego}$. Therefore, the constraints of the left and right road boundaries can be written as:
\begin{equation}\label{eq:ConstraintsBordi}
\pm y^{ego}_{c,\star}+\rho^{ego} \le \pm (\sum_{i=1}^{4} K_{L(i),R(i)}\cdot s^{n-i})\\
\end{equation}
\noindent with $\star = \{1,2\}$ and $K_{L(i)}$ and $K_{R(i)}\,$, the $4$ coefficients of the third order left and right polynomials respectively. Soft constraints are also implemented using an exponential function:
\begin{equation}\label{eq:ConstraintsBordi-soft}
C_{\{left,right\},\star}\!=\! e^{\,\kappa \big[\rho^{ego} \pm (y^{ego}_{c,\star}-(\sum_{i=1}^{n} K_{L(i),R(i)}\cdot s^{n-i}))\big]}\\
\end{equation}

The resulting penalty terms $C_{left,\star}$ and $C_{right,\star}$ are accounted for in the cost function.

\subsection{Cost Function}
The cost function to be minimized is in the form of Bolza problem. Thus, it includes a quadratic integral term and a quadratic terminal term:
\begin{multline}
J=\frac{1}{2}\int_{t_0}^{t_f} \big(\ ||\bm{h}(\tau) - \bm{h_{ref}}(\tau)||_Q^2 + ||\bm{u}(\tau)||_R^2\ \big)\ d\tau\  \\ +\  \frac{1}{2}||\bm{h}(t_f)-\bm{h_{ref}}(t_f)||_P^2
\end{multline}

\noindent where $||\bullet||_{Q,R,P}^2$ denotes the Euclidean norm weighted with diagonal matrices $Q$, $R$ or $P$. The input vector $\bm{u}(t)\in \mathbb{R}^2$ accounts for the control actions, while $\bm{h^{\top}}(t)=[\bm{x^{\top}}(t), {\bm{C^{soft\,\top}}}(t)]$ is a time-dependent vector that contains the states $\bm{x}(t)\in \mathbb{R}^8$ and the cost penalties $\bm{C^{soft}}(t)\in \mathbb{R}^{N_{soft}}$, and $\bm{h_{ref}}(t)$ is the corresponding reference vector.

The elements of matrix $Q$ that refer to the states are used to normalize the state error with respect to the maximum desired deviation $\Delta x_{i_{max}}=x_{i_{max}}\!-\!h_{ref,i}$:
\begin{equation}\label{eq:CostDelta_h}
Q_{i,i}=\frac{1}{\Delta x_{i_{max}}^2}\qquad \text{for}\qquad i=1,...,N_x.
\end{equation}

In this way, the cost variation associated to an error equal to $\Delta x_{i_{max}}$ is the same for all state variables. The remaining $N_{soft}$ elements of the weight matrix $Q$ are related to soft constraints. Recalling the definitions of soft constraints \eqref{eq:ConstraintsOst-DoppioCerchio-soft} and \eqref{eq:ConstraintsBordi-soft}, the values of the weights $Q_{i,i}$ for $i=N_x\!+\!1,...,N_{soft}$ represent the cost penalty associated to soft constraints with exponents equal to zero, i.e. the cost penalty associated to the activation of the corresponding hard constraints \eqref{eq:ConstraintsOst-DoppioCerchio-hard} and \eqref{eq:ConstraintsBordi}. This holds also for soft constraints that are not coupled with a hard constraint such as \eqref{eq:ConstraintsOst-Ellissoide-soft}.

The reference vector $\bm{h_{ref}}$ is, in general, time dependent. All reference values related to soft constraints are set to zero. The reference for $\xi$, $V_y$, $\omega$, $\delta$, $T_r$ and all reference values related to soft constraints are set to zero. Instead, the reference value of $y$ can be different from zero in case of roads with multiple lanes. Similarly, the reference value of $V_x$ is set to be equal to a desired speed profile while the reference value of the curvilinear position $s$ is set to be equal to the time integral of the velocity or to an arbitrary value if the associated weight in matrix $Q$ is null.

The weight matrix $R$ can be chosen in the same way as done for the matrix $Q$ considering that the references for both control actions is zero.

Finally, matrix $P$ can be, for example, obtained as the solution of Riccati equation for the infinite time LQR problem considering the system's equations linearized around the steady state condition $\bm{x}= \bm{0}$ and weight matrices $Q$ and $R$.

\section{Algorithm Implementation}\label{section3}

\subsection{Perceived and Simulated Spatial Horizons}\label{sec:SpatialHorizon}

While developing a trajectory planner for autonomous driving, it is important to consider the difference between the horizon \emph{perceived} by the sensors and the one \emph{simulated} within the NMPC. In particular, the former depends on the range of the sensors (such as cameras, lidars and radars) while the latter is defined by two parameters: the time span of the optimization horizon, which is generally fixed \textit{a priori}, and the vehicle velocity, which is not known beforehand being a consequence of the optimization process itself. It is clear that, for safety reasons, the simulated spatial horizon should be shorter than the perceived one, thus allowing to plan trajectories only within the ``known'' environment.

This issue can be addressed by transforming the OCP into a spatial dependent one as in \cite{frasch2013auto} and \cite{gao2012spatial}. As already pointed out, this approach implies strong assumptions on the velocity profile of the vehicle.

\begin{figure}
	\centering
	{\includegraphics[width=0.35\textwidth]{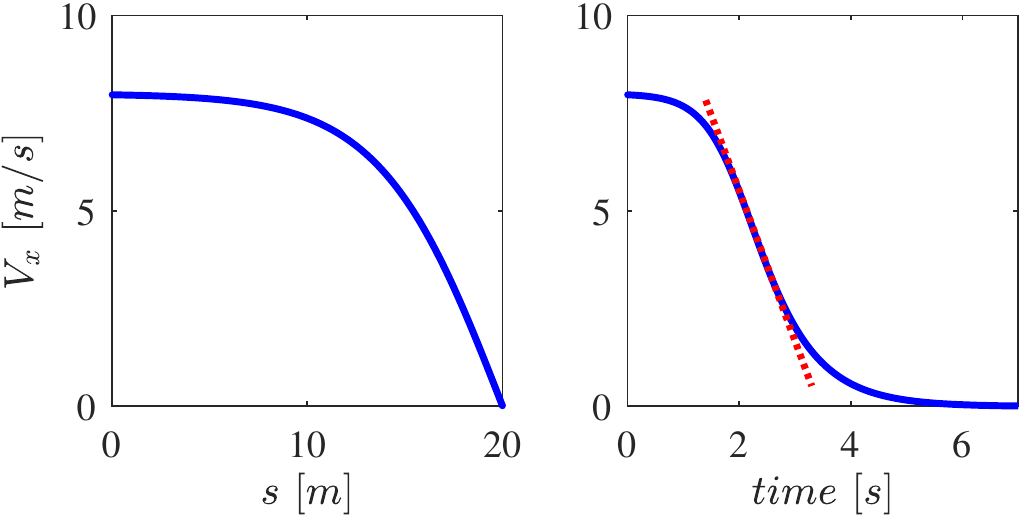}}
	\caption{Spatial and temporal velocity profiles.}
	\label{fig:TemporalVelProfile}
\end{figure}

In order not to switch to a spatial OCP, the simulated spatial horizon is limited by imposing an extra inequality constraint on the velocity $V_x$ as a function of the curvilinear abscissa $s$ as in \eqref{eq:ConstraintProfiloVel}. The velocity profile constraint enforces a deceleration that achieves zero velocity at the perceived spatial horizon limit.
\begin{equation}\label{eq:ConstraintProfiloVel}
V_x \leq V_{x_{ref}}\tanh\! \big(\kappa_{vel}(s-s_{max})\big)
\end{equation}
\noindent The reference velocity $V_{x_{ref}}$ is set by the behavioural layer as the maximum allowed speed considering road rules and estimated tyre-road friction coefficient. The parameter $\kappa_{vel}$ is used to limit the resulting maximum deceleration imposed by the velocity profile within comfort limits. In Fig.~\ref{fig:TemporalVelProfile} examples of spatial and temporal velocity profiles for $V_{x_{ref}}\!=\!8\ m/s$ and $s_{max}\!=\!20\ m$ are shown. Differently from a constant deceleration, this profile provides a much smoother deceleration, especially toward the end of the braking phase, reducing harshness and improving passengers' comfort.

A similar effect could have been achieved by imposing a constraint on the maximum distance $s$ at the final time $t_N$. However, exploiting the velocity profile constraints, a smoother deceleration is obtained.

\subsection{Driving Modes}
Two driving modes have been implemented: the DRIVE and the OVERTAKE modes. The DRIVE mode is used to drive along the road, keeping the current lane and reducing the vehicle's velocity if a slower preceding vehicle is detected. The OVERTAKE mode, instead, is used to overtake slower preceding vehicles if possible. The switching between these two modes (and eventually between other modes) is demanded to the behavioural layer that considers road rules, traffic conditions and visibility range. The switch is accomplished by gradually shifting the weighting matrices of one mode to the matrices of the other mode. In this way the homotopy between subsequent solutions is ensured and sharp discontinuities are avoided, resulting in a more natural and comfortable driving behaviour.

Regarding stop-and-go manoeuvres, these are managed by the DRIVE mode, applying a spatial velocity profile that is gradually shifted to impose zero velocity at the desired stopping point. An extra hard inequality constraint on the maximum distance $s$ can also be applied.
	
\subsection{Numerical Solution}

The resulting finite horizon OCP that has to be solved at each NMPC iteration is the following:
\begin{subequations}\label{eq:OCP}
	\begin{align}
	\underset{\bm{u}(t)}{\text{minimize }} &\frac{1}{2}\int_{t_0}^{t_f} \big(\ ||\bm{h}(\tau)-\bm{h_{ref}}(\tau)||_Q^2 + ||\bm{u}(\tau)||_R^2\ \big)\ d\tau\  \notag \\
	&\qquad \qquad+\  \frac{1}{2}||\bm{h}(t_f)-\bm{h_{ref}}(t_f)||_P^2\\
	\text{subject to:}\ &\dot{\bm{x}}(t)=f(\bm{x}(t),\bm{u}(t))\\
	&\bm{x}(t_0)=\bm{x_0}\\
	&\bm{x_{min}}(t)\leq \bm{x}(t) \leq \bm{x_{max}}(t)\\
	&\bm{u_{min}}(t)\leq \bm{u}(t) \leq \bm{u_{max}}(t)\\
	&\bm{g}(t)\leq \bm{0}
	\end{align}
\end{subequations}

The OCP is iteratively solved in a MPC framework using the open source ACADO toolkit software that implements Bock's multiple shooting method \cite{bock1984multiple} coupled with Real-Time Iteration (RTI) scheme~\cite{diehl2002real, diehl2005real, diehl2007stabilizing}.

The RTI scheme is based on a sequential quadratic programming (SQP) method that performs only one iteration per sampling time, reinitializing the trajectory by shifting the old prediction. In particular, each RTI iteration is divided into a preparation and a feedback phase. The preparation phase is performed before the new sample is available and takes care of computationally expensive tasks, such as sensitivities generation and matrix condensation. The feedback phase is performed, as soon as the new sample is available, by solving the updated small-scale parametric QP problem. In this case, the small-scale QP problem has been solved with the QP solver qpOASES \cite{ferreau2014qpoases}, an open source C++ implementation of an online active-set strategy based on a parametric quadratic programming approach \cite{best1996algorithm}.

The ACADO Code Generation tool produces a self-contained C-code. The computational speed is increased exploiting automatic differentiation and avoiding dynamic memory allocation by hard-coding all problem dimensions~\cite{houska2011acado}.

\subsection{Improved Robustness}\label{sec:ImprovedRobustness}
The presence of hard constraints and the nonlinearity of the problem may induce an infeasibility, causing a solver error, if an obstacle suddenly appears within the simulated spatial horizon, e.g. a pedestrian suddenly crosses the street or another vehicle performs a hazardous overtaking manoeuvre. For these reasons, it is necessary to set up a backup policy that allows, if possible, to find a solution to the planning task, while guaranteeing feasibility.

A possible solution could be to reinitialize all the solver steps at each iteration with the current state estimation. This method can be successfully applied for low speed manoeuvres with short simulated spatial horizons in particularly crowded environments, such as a restart after a traffic stop light. Unfortunately, as the spatial horizon and the longitudinal velocity increase ($V_x \gtrapprox 2\ m/s$), this method would require too many QP iterations to converge to a locally optimal solution.

An alternative way is to have two or more identical sub-planners with different simulated spatial horizon lengths that work in parallel. Let's consider sub-planners $A$ and $B$ having decreasing horizon lengths: sub-planner $A$ has the longest simulated spatial horizon; sub-planner $B$, instead, has a simulated spatial horizon length that corresponds to the minimum stopping distance calculated considering the current vehicle's velocity and maximum admissible deceleration. The sub-planners are run in parallel, with a limited number of working set recalculations to account for the limited available computation time of the overall optimization. The solution adopted is the one given by the sub-planner with the longer spatial horizon that obtained a feasible trajectory.
	
\section{Simulation Results}\label{section4}
The NMPC described in previous sections is implemented considering a time horizon length of $3\,s$ discretized in $N=60$ steps (thus $\Delta t=0.05s$). The temporal horizon length is a design parameter that should be chosen considering the range of the sensors and the stopping distance of the vehicle (a function of tyre-road friction coefficient and velocity). 
The forward integration of the system's dynamics is accomplished by second order implicit Runge-Kutta method and two full sequential quadratic programming (SQP) iterations are performed per NMPC step. Thus, the vehicle's inputs are updated with a frequency of $20\ Hz$, applying the first of the $60$ calculated optimal control actions.

All closed-loop simulations are performed considering a detailed 18 dofs CarMaker model. Ego-vehicle state and obstacles positions and velocities are directly computed by the simulator and assumed as known. In \cite{Bersani2021} a real implementation of and estimator for these quantities based on a kalman filter algorithm that combines lidars and radars data is presented. Simulations are performed on a laptop featuring an Intel i7-3610QM CPU and 8 GB of RAM.

In the following, two relevant driving scenarios are considered.

\subsection{Overtaking of a single moving obstacle}
In this first scenario the ego-vehicle is required to overtake a moving obstacle positioned at a distance $\Delta s=25\,m$, in the same lane as the ego-vehicle, with constant longitudinal velocity $V_{s_{obs}}=10\ m/s$ and zero lateral velocity. The reference longitudinal velocity of ego-vehicle is set to $V_{x_{ref}}=13\ m/s$ and a straight road is considered for the sake of representation clarity.

\begin{figure}
	\centering
	{\includegraphics[width=0.47\textwidth]{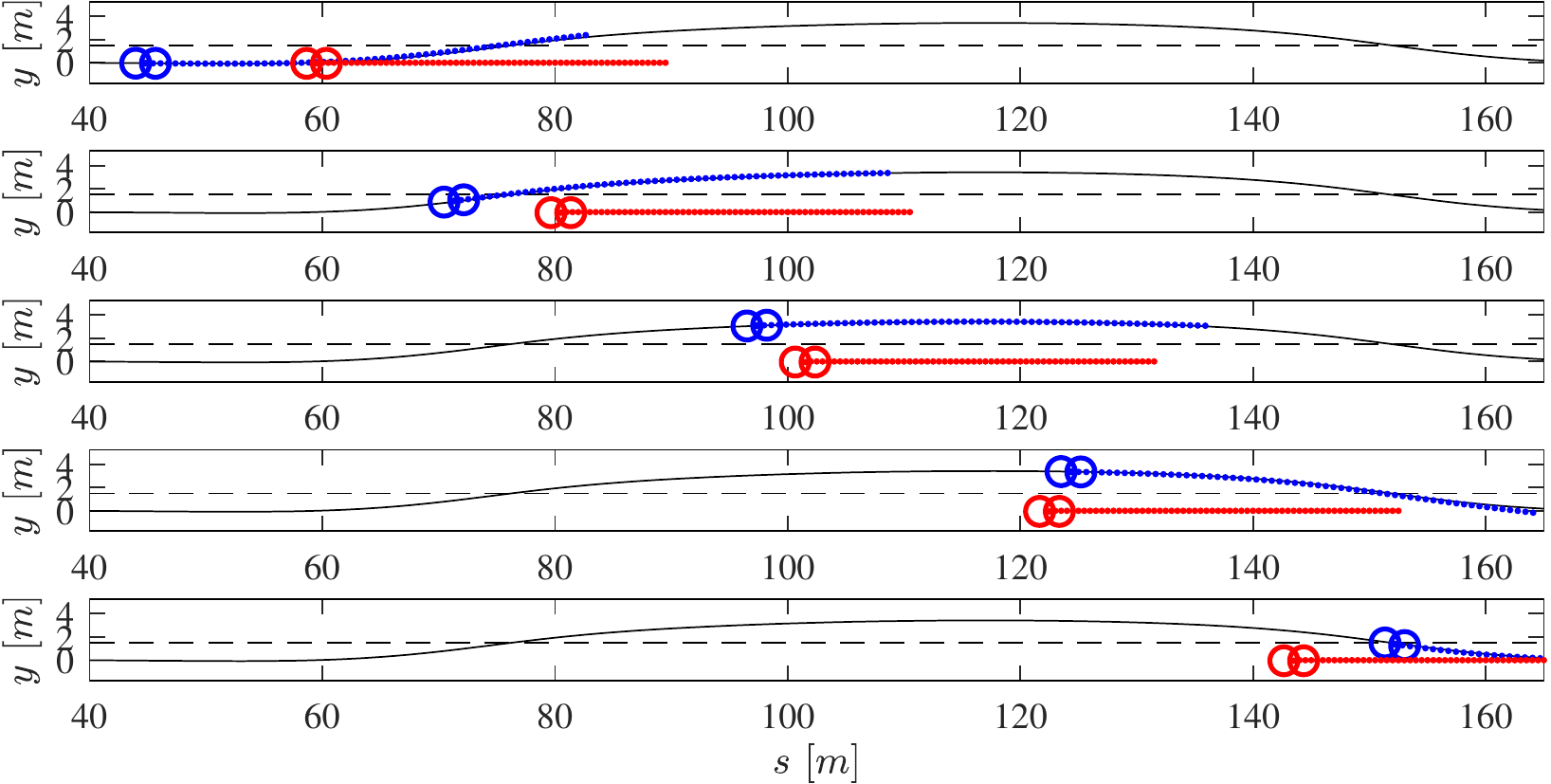}}
	\caption{Ego-vehicle (blue) and obstacle (red) trajectories during the overtaking manoeuvre.}
	\label{fig:Traiettoria_SMO_O_8_5}
\end{figure}

The weight matrix of the OVERTAKE mode and a sufficiently long perceived spatial horizon allow a safe overtaking manoeuvre, during which the safety distance from the obstacle is always maintained. The ego-vehicle and obstacle trajectories, together with the predictions along the optimization horizon, are displayed in Fig. \ref{fig:Traiettoria_SMO_O_8_5}.

\begin{figure}
	\centering
	{\includegraphics[width=0.32\textwidth]{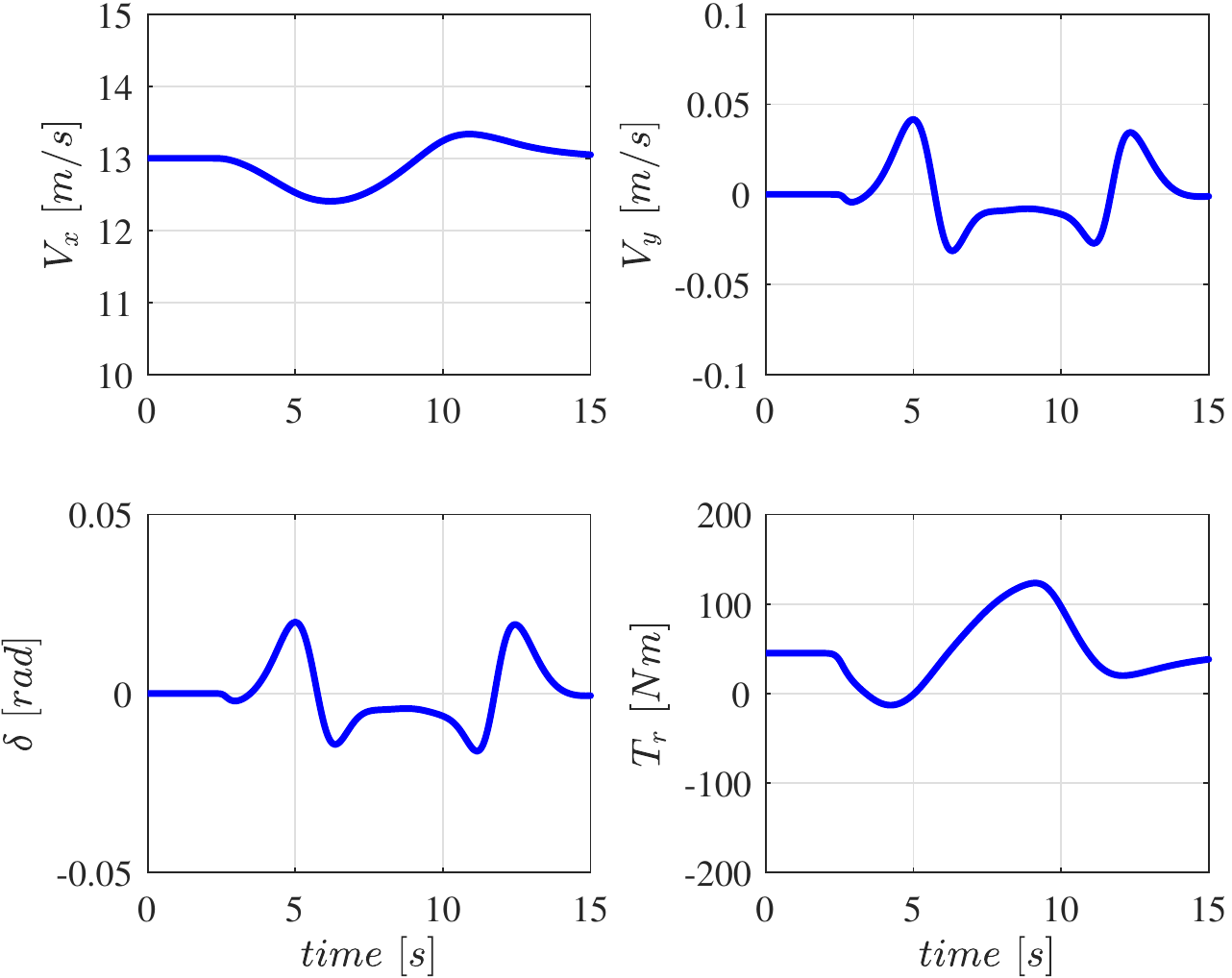}}
	\caption{Longitudinal velocity $V_x$, lateral velocity $V_y$, steering angle at the wheels $\delta$ and rear axle torque $T_r$ during the overtake manoeuvre.}
	\label{fig:Stato_Ridotto_SMO_O_8_5}
\end{figure}

The velocities $V_x$ and $V_y$, the steering angle $\delta$ and the applied torque $T_r$ as functions of time are presented in Fig.~\ref{fig:Stato_Ridotto_SMO_O_8_5}. It should be noted that the velocity of the ego-vehicle decreases by approximately $0.5\, m/s$ in the first part of the manoeuvre and successively increases to the reference value with a small overshoot. The lateral displacement is increased up to $y=3.4\ m$ to keep the appropriate safety distance from the obstacle during the overtaking manoeuvre.

The computational time required to solve each OCP is, on average, lower than $8 \, ms$. Therefore, the real-time feasibility of the proposed trajectory planner can be stated.

\subsection{Car suddenly exiting from a blind spot}
\begin{figure}
	\centering
	{\includegraphics[width=0.28\textwidth]{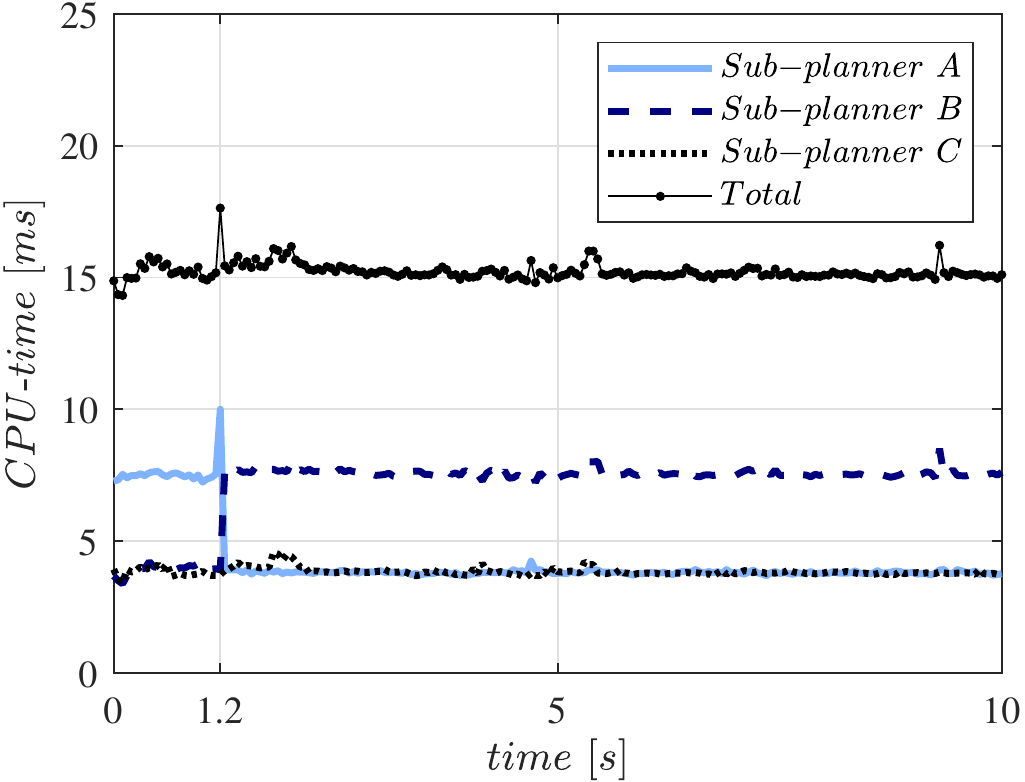}}
	\caption{sub-planner $A$, sub-planner $B$, sub-planner $C$ and total trajectory planner CPU-time.}
	\label{fig:time_TRE_solutori}
\end{figure}

The parallel sub-planners structure proposed in Section \ref{sec:ImprovedRobustness} is now assessed considering three sub-planners, $A$, $B$ and $C$, with decreasing spatial horizon lengths. The simulation test is performed on a straight road along which the ego-vehicle is running at a speed equal to $V_{x}=8\ m/s$. An obstacle suddenly appears from the right side of the road at a distance $\Delta s=20\ m$. The obstacle travels at $3\ m/s$ and enters the ego-vehicle lane without observing the right of way prescribed by the yield sign. Due to the presence of a building, the field of view of the ego-vehicle is obstructed and the other vehicle is detected only at $t=1.2\ s$. Hence, an infeasibility may occur in sub-planners characterized by long spatial horizons. This is shown in Fig.\ref{fig:time_TRE_solutori}, where a sudden growth in computational time affects sub-planner $A$: in these conditions, sub-planners $B$ and $C$ can provide feasible manoeuvres. In particular, at time $t=1.2\ s$, sub-planner $A$ cannot find a feasible solution within the prescribed number of iterations or CPU-time. Thus, the solution of sub-planner $B$ is adopted. At the next iteration, sub-planned $B$ becomes the leading planner and its simulated spatial horizon is extended to the perceived spatial horizon length while sub-planner $A$ is reinitialized with a simulated spatial horizon equal to that of $B$ at the previous step.

\begin{figure}
	\centering
	{\includegraphics[width=0.32\textwidth]{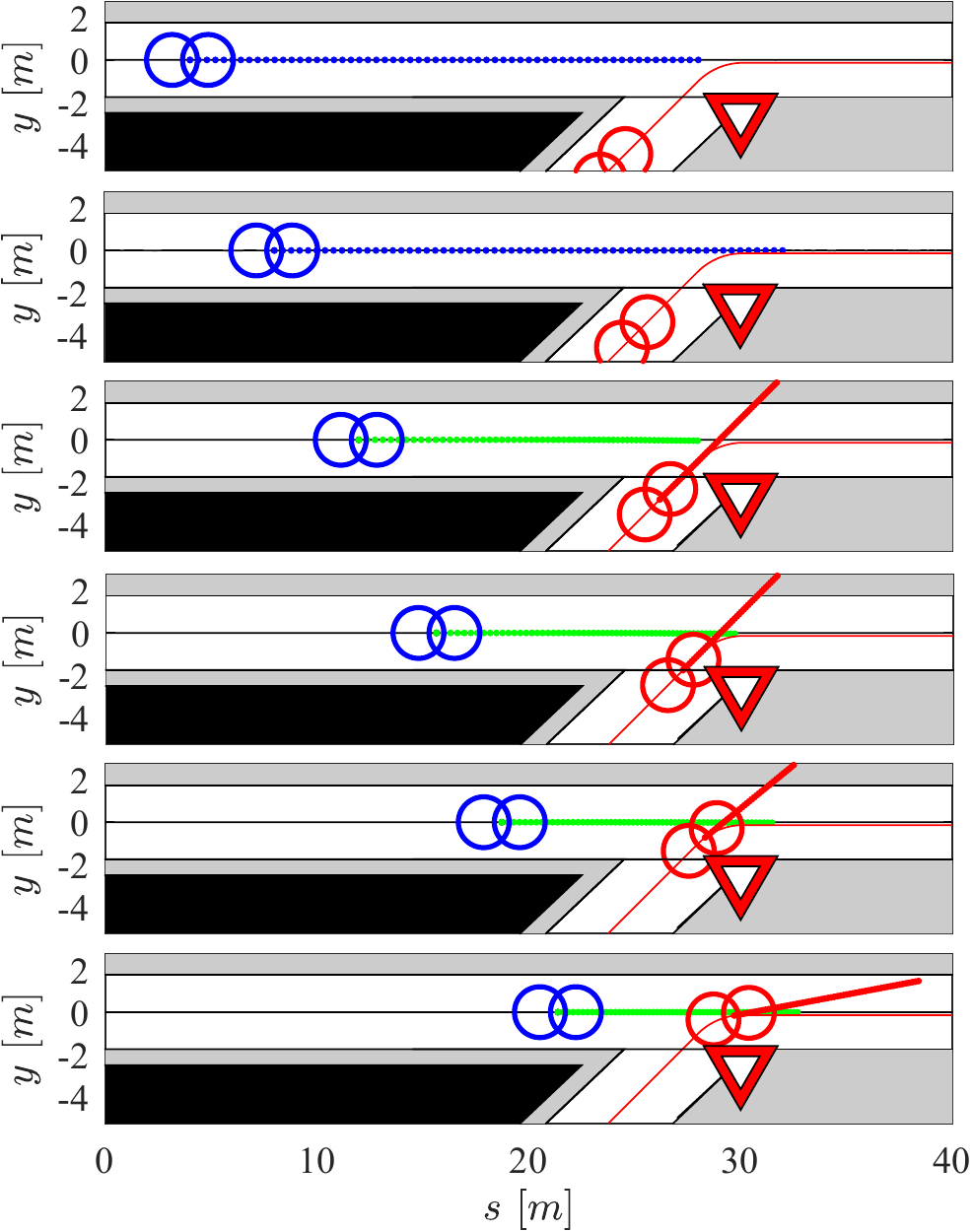}}
	\caption{Ego-vehicle and obstacle trajectories. The obstacle (red) suddenly appears on the side of the ego-vehicle (blue) due to an obstruction of its field of view.
	}
	\label{fig:Traiettoria_3_solutori}
\end{figure}

The ego-vehicle and obstacle trajectories, together with their prediction along the optimization horizon, are displayed in Fig.~\ref{fig:Traiettoria_3_solutori}. Note that, as explained in Section~\ref{section2}, the obstacle trajectory is approximated at each step with constant velocities in both longitudinal and lateral directions.

\begin{figure}
	\centering
	{\includegraphics[width=0.32\textwidth]{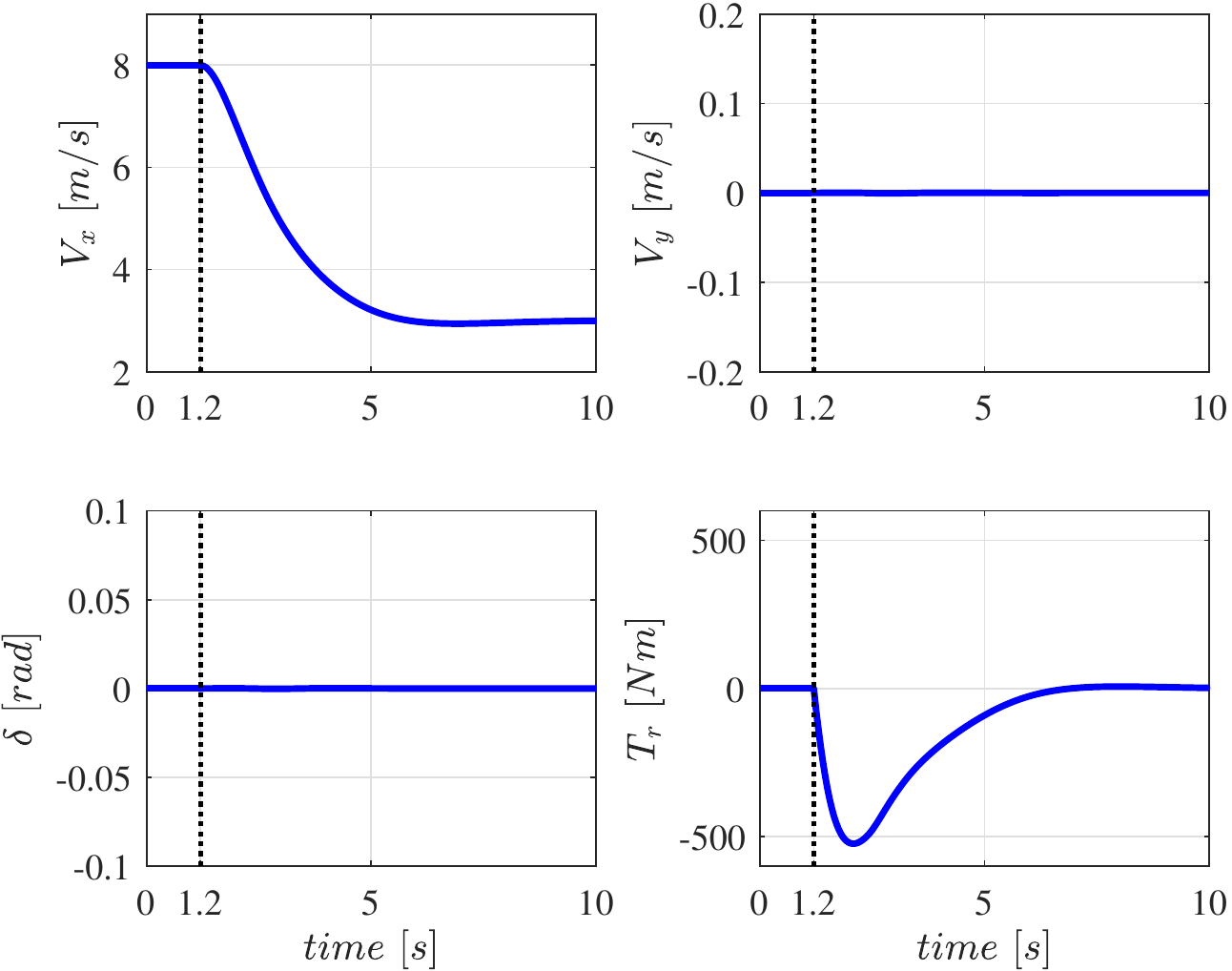}}
	\caption{Longitudinal velocity $V_x$, lateral velocity $V_y$, steering angle at the wheels $\delta$ and rear axle torque $T_r$ during the emergency manoeuvre. At $t=1.2\ s$ sub-planner $A$ cannot find a feasible solution. Thus, the solution provided by sub-planner $B$ is used instead.}
	\label{fig:Stato_Ridotto_TRE_solutori}
\end{figure}

Fig.~\ref{fig:Stato_Ridotto_TRE_solutori}, instead, shows the velocities $V_x$ and $V_y$, the steering angle $\delta$ and the applied torque $T_r$ during the braking manoeuvre. The longitudinal velocity is reduced to match the speed of the merging vehicle, generating a longitudinal deceleration of approximately $6.5\ m/s^2$, which is still within the adhesion limit of the tyres.

The trajectory planner with three parallel sub-planners shows improved robustness and guarantees feasibility in a broader range of scenarios, while achieving an overall smooth manoeuvre that does not present discontinuities.

Aiming at reducing the computational burden of this configuration, 
only the sub-planner with the longest simulated spatial horizon performs two full SQP iterations, while the other two implement only one SQP iteration. The loop times required by the three sub-planners are shown in Fig.~\ref{fig:time_TRE_solutori}. It is possible to notice that, at $t\!=\!1.2\ s$, sub-planner $A$ reaches the maximum allowed CPU-time of $10\ ms$. Thus, the solution calculated by sub-planner $B$ is considered while sub-planner $A$ is reinitialized. In the next step, sub-planner $B$ takes the role of sub-planner $A$, acquiring the longer spatial horizon and performing two SQP iterations. Even without considering possible computation parallelization, the total computational time, even in this configuration, is lower than $20\ ms$, thus well below the total available time of $50\ ms$.

\section{Conclusions}\label{section5}
In this paper a NMPC trajectory planner for autonomous vehicles based on a direct approach has been presented. Thanks to the proposed modified slip calculation, the dynamic single-track vehicle model can be employed at both high and low velocities. This allows to have one single model that presents a kinematic-like behaviour at low velocities while keeping the standard dynamic single-track vehicle model behaviour at higher speeds. Obstacles' uncertainties have been taken into account through the implementation of coupled soft and hard constraints. The correlation between the simulated and perceived spatial horizon lengths has been enforced with a spatial dependent velocity profile that has been imposed as an extra inequality constraint. The numerical solution is carried out using ACADO toolkit, coupled with the QP solver qpOASES. The trajectory planner performances have been checked in simulation, applying the controller to a nonlinear multibody model in CarMaker environment. The results coming from the simulations carried on two significant driving scenarios have been reported to highlight the effectiveness of the calculated trajectories. The analysis of the computational times has confirmed that the proposed trajectory planner can be implemented within the real-time control routine of an autonomous vehicle.

\bibliographystyle{IEEEtran}
\bibliography{bibliografia} 

\begin{thebibliography}{10}
\providecommand{\url}[1]{#1}
\csname url@samestyle\endcsname
\providecommand{\newblock}{\relax}
\providecommand{\bibinfo}[2]{#2}
\providecommand{\BIBentrySTDinterwordspacing}{\spaceskip=0pt\relax}
\providecommand{\BIBentryALTinterwordstretchfactor}{4}
\providecommand{\BIBentryALTinterwordspacing}{\spaceskip=\fontdimen2\font plus
\BIBentryALTinterwordstretchfactor\fontdimen3\font minus
  \fontdimen4\font\relax}
\providecommand{\BIBforeignlanguage}[2]{{%
\expandafter\ifx\csname l@#1\endcsname\relax
\typeout{** WARNING: IEEEtran.bst: No hyphenation pattern has been}%
\typeout{** loaded for the language `#1'. Using the pattern for}%
\typeout{** the default language instead.}%
\else
\language=\csname l@#1\endcsname
\fi
#2}}
\providecommand{\BIBdecl}{\relax}
\BIBdecl

\bibitem{harper2016estimating}
C.~D. Harper, C.~T. Hendrickson, S.~Mangones, and C.~Samaras, ``Estimating
  potential increases in travel with autonomous vehicles for the non-driving,
  elderly and people with travel-restrictive medical conditions,''
  \emph{Transportation research part C: emerging technologies}, vol.~72, pp.
  1--9, 2016.

\bibitem{greenblatt2015automated}
J.~B. Greenblatt and S.~Shaheen, ``Automated vehicles, on-demand mobility, and
  environmental impacts,'' \emph{Current sustainable/renewable energy reports},
  vol.~2, no.~3, pp. 74--81, 2015.

\bibitem{paden2016survey}
B.~Paden, M.~{\v{C}}{\'a}p, S.~Z. Yong, D.~Yershov, and E.~Frazzoli, ``A survey
  of motion planning and control techniques for self-driving urban vehicles,''
  \emph{IEEE Transactions on intelligent vehicles}, vol.~1, no.~1, pp. 33--55,
  2016.

\bibitem{li2017development}
X.~Li, Z.~Sun, D.~Cao, D.~Liu, and H.~He, ``Development of a new integrated
  local trajectory planning and tracking control framework for autonomous
  ground vehicles,'' \emph{Mechanical Systems and Signal Processing}, vol.~87,
  pp. 118--137, 2017.

\bibitem{khatib1985real}
O.~Khatib, ``Real-time obstacle avoidance for manipulators and mobile robots,''
  in \emph{Proceedings. 1985 IEEE International Conference on Robotics and
  Automation}, vol.~2.\hskip 1em plus 0.5em minus 0.4em\relax IEEE, 1985, pp.
  500--505.

\bibitem{borenstein1989real}
J.~Borenstein and Y.~Koren, ``Real-time obstacle avoidance for fast mobile
  robots,'' \emph{IEEE Transactions on systems, Man, and Cybernetics}, vol.~19,
  no.~5, pp. 1179--1187, 1989.

\bibitem{Borenstein1991}
------, ``The vector field histogram-fast obstacle avoidance for mobile
  robots,'' \emph{IEEE transactions on robotics and automation}, vol.~7, no.~3,
  pp. 278--288, 1991.

\bibitem{koren1991potential}
Y.~Koren and J.~Borenstein, ``Potential field methods and their inherent
  limitations for mobile robot navigation,'' in \emph{Robotics and Automation,
  1991. Proceedings., 1991 IEEE International Conference on}.\hskip 1em plus
  0.5em minus 0.4em\relax IEEE, 1991, pp. 1398--1404.

\bibitem{judd2001spline}
K.~Judd and T.~McLain, ``Spline based path planning for unmanned air
  vehicles,'' in \emph{AIAA Guidance, Navigation, and Control Conference and
  Exhibit}, 2001, p. 4238.

\bibitem{fleury1995primitives}
S.~Fleury, P.~Soueres, J.-P. Laumond, and R.~Chatila, ``Primitives for
  smoothing mobile robot trajectories,'' \emph{IEEE transactions on robotics
  and automation}, vol.~11, no.~3, pp. 441--448, 1995.

\bibitem{lavalle1998rapidly}
S.~M. LaValle, ``Rapidly-exploring random trees: A new tool for path
  planning,'' 1998.

\bibitem{karaman2010optimal}
S.~Karaman and E.~Frazzoli, ``Optimal kinodynamic motion planning using
  incremental sampling-based methods,'' in \emph{Decision and Control (CDC),
  2010 49th IEEE Conference on}.\hskip 1em plus 0.5em minus 0.4em\relax IEEE,
  2010, pp. 7681--7687.

\bibitem{berntorp2017path}
K.~Berntorp, ``Path planning and integrated collision avoidance for autonomous
  vehicles,'' in \emph{2017 American Control Conference (ACC)}.\hskip 1em plus
  0.5em minus 0.4em\relax IEEE, 2017, pp. 4023--4028.

\bibitem{pepy2006path}
R.~Pepy, A.~Lambert, and H.~Mounier, ``Path planning using a dynamic vehicle
  model,'' in \emph{2006 2nd International Conference on Information \&
  Communication Technologies}, vol.~1.\hskip 1em plus 0.5em minus 0.4em\relax
  IEEE, 2006, pp. 781--786.

\bibitem{nagy2001trajectory}
B.~Nagy and A.~Kelly, ``Trajectory generation for car-like robots using cubic
  curvature polynomials,'' \emph{Field and Service Robots}, vol.~11, 2001.

\bibitem{chen2014quartic}
C.~Chen, Y.~He, C.~Bu, J.~Han, and X.~Zhang, ``Quartic b{\'e}zier curve based
  trajectory generation for autonomous vehicles with curvature and velocity
  constraints,'' in \emph{2014 IEEE International Conference on Robotics and
  Automation (ICRA)}.\hskip 1em plus 0.5em minus 0.4em\relax IEEE, 2014, pp.
  6108--6113.

\bibitem{alia2015local}
C.~Alia, T.~Gilles, T.~Reine, and C.~Ali, ``Local trajectory planning and
  tracking of autonomous vehicles, using clothoid tentacles method,'' in
  \emph{2015 IEEE Intelligent Vehicles Symposium (IV)}.\hskip 1em plus 0.5em
  minus 0.4em\relax IEEE, 2015, pp. 674--679.

\bibitem{de2009flatness}
J.~De~Don{\'a}, F.~Suryawan, M.~Seron, and J.~L{\'e}vine, ``A flatness-based
  iterative method for reference trajectory generation in constrained nmpc,''
  in \emph{Nonlinear Model Predictive Control}.\hskip 1em plus 0.5em minus
  0.4em\relax Springer, 2009, pp. 325--333.

\bibitem{qian2016motion}
X.~Qian, I.~Navarro, A.~de~La~Fortelle, and F.~Moutarde, ``Motion planning for
  urban autonomous driving using b{\'e}zier curves and mpc,'' in \emph{2016
  IEEE 19th International Conference on Intelligent Transportation Systems
  (ITSC)}.\hskip 1em plus 0.5em minus 0.4em\relax Ieee, 2016, pp. 826--833.

\bibitem{falcone2007predictive}
P.~Falcone, F.~Borrelli, J.~Asgari, H.~E. Tseng, and D.~Hrovat, ``Predictive
  active steering control for autonomous vehicle systems,'' \emph{IEEE
  Transactions on control systems technology}, vol.~15, no.~3, 2007.

\bibitem{falcone2008low}
P.~Falcone, F.~Borrelli, J.~Asgari, H.~Tseng, and D.~Hrovat, ``Low complexity
  mpc schemes for integrated vehicle dynamics control problems,'' in \emph{9th
  international symposium on advanced vehicle control}, 2008.

\bibitem{liniger2015optimization}
A.~Liniger, A.~Domahidi, and M.~Morari, ``Optimization-based autonomous racing
  of 1: 43 scale rc cars,'' \emph{Optimal Control Applications and Methods},
  vol.~36, no.~5, pp. 628--647, 2015.

\bibitem{frasch2013auto}
J.~V. Frasch, A.~Gray, M.~Zanon, H.~J. Ferreau, S.~Sager, F.~Borrelli, and
  M.~Diehl, ``An auto-generated nonlinear mpc algorithm for real-time obstacle
  avoidance of ground vehicles,'' in \emph{Control Conference (ECC), 2013
  European}.\hskip 1em plus 0.5em minus 0.4em\relax IEEE, 2013, pp. 4136--4141.

\bibitem{gray2012predictive}
A.~Gray, Y.~Gao, T.~Lin, J.~K. Hedrick, H.~E. Tseng, and F.~Borrelli,
  ``Predictive control for agile semi-autonomous ground vehicles using motion
  primitives,'' in \emph{American Control Conference (ACC), 2012}.\hskip 1em
  plus 0.5em minus 0.4em\relax IEEE, 2012, pp. 4239--4244.

\bibitem{gao2012spatial}
Y.~Gao, A.~Gray, J.~V. Frasch, T.~Lin, E.~Tseng, J.~K. Hedrick, and
  F.~Borrelli, ``Spatial predictive control for agile semi-autonomous ground
  vehicles,'' in \emph{Proceedings of the 11th international symposium on
  advanced vehicle control}, 2012.

\bibitem{gutjahr2017lateral}
B.~Gutjahr, L.~Gr{\"o}ll, and M.~Werling, ``Lateral vehicle trajectory
  optimization using constrained linear time-varying mpc,'' \emph{IEEE
  Transactions on Intelligent Transportation Systems}, vol.~18, no.~6, 2017.

\bibitem{kong2015kinematic}
J.~Kong, M.~Pfeiffer, G.~Schildbach, and F.~Borrelli, ``Kinematic and dynamic
  vehicle models for autonomous driving control design.'' in \emph{Intelligent
  Vehicles Symposium}, 2015, pp. 1094--1099.

\bibitem{inghilterra2018}
G.~Inghilterra, S.~Arrigoni, F.~Braghin, and F.~Cheli, ``Firefly
  algorithm-based nonlinear mpc trajectory planner for autonomous driving,''
  \emph{2018 International Conference of Electrical and Electronic Technologies
  for Automotive}, 2018.

\bibitem{trabalzini2019}
S.~Arrigoni, E.~Trabalzini, M.~Bersani, F.~Braghin, and F.~Cheli, ``Non-linear
  mpc motion planner for autonomous vehicles based on accelerated particle
  swarm ptimization algorithm,'' \emph{2019 International Conference of
  Electrical and Electronic Technologies for Automotive}, 2019.

\bibitem{Arrigoni2021mpc}
\BIBentryALTinterwordspacing
S.~Arrigoni, F.~Braghin, and F.~Cheli, ``Mpc path-planner for autonomous
  driving solved by genetic algorithm technique,'' 2021. [Online]. Available:
  \url{https://arxiv.org/abs/2102.01211}
\BIBentrySTDinterwordspacing

\bibitem{pacejka1992magic}
H.~B. Pacejka and E.~Bakker, ``The magic formula tyre model,'' \emph{Vehicle
  system dynamics}, vol.~21, no.~S1, pp. 1--18, 1992.

\bibitem{houska2011acado}
B.~Houska, H.~J. Ferreau, and M.~Diehl, ``Acado toolkit—an open-source
  framework for automatic control and dynamic optimization,'' \emph{Optimal
  Control Applications and Methods}, vol.~32, no.~3, pp. 298--312, 2011.

\bibitem{ferreau2014qpoases}
H.~J. Ferreau, C.~Kirches, A.~Potschka, H.~G. Bock, and M.~Diehl, ``qpoases: A
  parametric active-set algorithm for quadratic programming,''
  \emph{Mathematical Programming Computation}, vol.~6, no.~4, 2014.

\bibitem{carmaker2014ipg}
``Carmaker 7.0.2,'' \emph{IPG Automotive GmbH, Karlsruhe, Germany,
  https://ipg-automotive.com/}.

\bibitem{cinesi_curva}
K.~{Jo}, M.~{Lee}, J.~{Kim}, and M.~{Sunwoo}, ``Tracking and behavior reasoning
  of moving vehicles based on roadway geometry constraints,'' \emph{IEEE
  Transactions on Intelligent Transportation Systems}, vol.~18, no.~2, pp.
  460--476, Feb 2017.

\bibitem{Mentasti2019}
S.~Mentasti and M.~Matteucci, ``Multi-layer occupancy grid mapping for
  autonomous vehicles navigation,'' in \emph{2019 International Conference of
  Electrical and Electronic Technologies for Automotive}.\hskip 1em plus 0.5em
  minus 0.4em\relax IEEE, 2019.

\bibitem{kocic2018sensors}
J.~Koci{\'c}, N.~Jovi{\v{c}}i{\'c}, and V.~Drndarevi{\'c}, ``Sensors and sensor
  fusion in autonomous vehicles,'' in \emph{2018 26th Telecommunications Forum
  (TELFOR)}.\hskip 1em plus 0.5em minus 0.4em\relax IEEE, 2018, pp. 420--425.

\bibitem{braghin2009dinamica}
F.~Braghin, F.~Cheli, E.~Leo, S.~Melzi, and E.~Sabbioni, ``Dinamica
  dell'autoveicolo,'' 2009.

\bibitem{shim2007understanding}
T.~Shim and C.~Ghike, ``Understanding the limitations of different vehicle
  models for roll dynamics studies,'' \emph{Vehicle system dynamics}, vol.~45,
  no.~3, pp. 191--216, 2007.

\bibitem{rajamani2011vehicle}
R.~Rajamani, \emph{Vehicle dynamics and control}.\hskip 1em plus 0.5em minus
  0.4em\relax Springer Science \& Business Media, 2011.

\bibitem{NILSSON1984}
\BIBentryALTinterwordspacing
N.~J. NILSSON, ``Shakey the robot,'' \emph{Technical Report}, no. 323, 1984.
  [Online]. Available: \url{https://ci.nii.ac.jp/naid/10004211424/en/}
\BIBentrySTDinterwordspacing

\bibitem{micaelli1993trajectory}
A.~Micaelli and C.~Samson, ``Trajectory tracking for unicycle-type and
  two-steering-wheels mobile robots,'' Ph.D. dissertation, INRIA, 1993.

\bibitem{brach2000modeling}
R.~M. Brach and R.~M. Brach, ``Modeling combined braking and steering tire
  forces,'' SAE Technical Paper, Tech. Rep., 2000.

\bibitem{brach2011tire}
R.~Brach and M.~Brach, ``The tire-force ellipse (friction ellipse) and tire
  characteristics,'' SAE Technical Paper, Tech. Rep., 2011.

\bibitem{mastinu2014road}
G.~Mastinu and M.~Ploechl, \emph{Road and off-road vehicle system dynamics
  handbook}.\hskip 1em plus 0.5em minus 0.4em\relax CRC Press, 2014.

\bibitem{vignati2018autonomous}
M.~Vignati, D.~Tarsitano, M.~Bersani, and F.~Cheli, ``Autonomous steer
  actuation for an urban quadricycle,'' in \emph{2018 International Conference
  of Electrical and Electronic Technologies for Automotive}.\hskip 1em plus
  0.5em minus 0.4em\relax IEEE, 2018.

\bibitem{vignati2018transform}
M.~Vignati, D.~Tarsitano, and F.~Cheli, ``On how to transform a commercial
  electric quadricycle into a full autonomously actuated vehicle,'' in
  \emph{14th Intenational Symposium on Advanced Vehicle Control (AVEC'18)},
  2018, pp. 1--7.

\bibitem{bock1984multiple}
H.~G. Bock and K.-J. Plitt, ``A multiple shooting algorithm for direct solution
  of optimal control problems,'' \emph{IFAC Proceedings Volumes}, vol.~17,
  no.~2, pp. 1603--1608, 1984.

\bibitem{diehl2002real}
M.~Diehl, H.~G. Bock, J.~P. Schl{\"o}der, R.~Findeisen, Z.~Nagy, and
  F.~Allg{\"o}wer, ``Real-time optimization and nonlinear model predictive
  control of processes governed by differential-algebraic equations,''
  \emph{Journal of Process Control}, vol.~12, no.~4, pp. 577--585, 2002.

\bibitem{diehl2005real}
M.~Diehl, H.~G. Bock, and J.~P. Schl{\"o}der, ``A real-time iteration scheme
  for nonlinear optimization in optimal feedback control,'' \emph{SIAM Journal
  on control and optimization}, vol.~43, no.~5, pp. 1714--1736, 2005.

\bibitem{diehl2007stabilizing}
M.~Diehl, R.~Findeisen, and F.~Allg{\"o}wer, ``A stabilizing real-time
  implementation of nonlinear model predictive control,'' in \emph{Real-Time
  PDE-Constrained Optimization}.\hskip 1em plus 0.5em minus 0.4em\relax SIAM,
  2007, pp. 25--52.

\bibitem{best1996algorithm}
M.~J. Best, ``An algorithm for the solution of the parametric quadratic
  programming problem,'' in \emph{Applied mathematics and parallel
  computing}.\hskip 1em plus 0.5em minus 0.4em\relax Springer, 1996, pp.
  57--76.

\bibitem{Bersani2021}
\BIBentryALTinterwordspacing
M.~Bersani, S.~Mentasti, P.~Dahal, S.~Arrigoni, M.~Vignati, F.~Cheli, and
  M.~Matteucci, ``An integrated algorithm for ego-vehicle and obstacles state
  estimation for autonomous driving,'' \emph{Robotics and Autonomous Systems},
  vol. 139, p. 103662, 2021. [Online]. Available:
  \url{https://www.sciencedirect.com/science/article/pii/S0921889020305029}
\BIBentrySTDinterwordspacing

\end{thebibliography}

\end{document}